\documentclass[10pt,letterpaper]{article}
\usepackage[letterpaper, margin=1in]{geometry}

\usepackage{times}
\usepackage{epsfig}
\usepackage{amsmath}
\usepackage{amssymb}
\usepackage{booktabs}
\usepackage{mathtools}
\usepackage{multirow}
\usepackage{colortbl}
\usepackage{xcolor}
\usepackage{pifont} % For checkmarks
\usepackage{epigraph}
\usepackage{lettrine}
\usepackage{lettrine,Zallman}
\usepackage{algorithm}      % algorithm
\usepackage{algorithmic}    %
\usepackage[misc]{ifsym}
\usepackage{graphicx}       % graphics
\usepackage{subfigure}
\usepackage{multirow}
\usepackage{overpic}
\usepackage{enumitem}
\usepackage{caption}
\usepackage{subcaption}

\usepackage{hyperref} %

\hypersetup{
    colorlinks=true,    % 
    citecolor=blue,     % 
    linkcolor=blue,     % 
    urlcolor=blue       % 
}

\definecolor{shapecolor}{rgb}{0.1,0.45,0.8}
\definecolor{lblue}{rgb}{0.1,0.45,0.8}
\definecolor{lgreen}{rgb}{0.18,0.58,0.18}

\newcommand{\modelname}{\textbf{CliffordNet}}
\newcommand{\cmark}{\ding{51}}
\newcommand{\xmark}{\ding{55}}
\definecolor{highlight}{HTML}{E6F4EA}

\title{CliffordNet: All You Need is Geometric Algebra}

\author{
Zhongping Ji\\
{}
}
\date{}
\begin{document}

\maketitle

% ----------------------------------------------------------------------
% ABSTRACT
% ----------------------------------------------------------------------

\begin{abstract}
Modern computer vision architectures, from CNNs to Transformers, predominantly rely on the stacking of heuristic modules: spatial mixers (Attention/Conv) followed by channel mixers (FFNs). In this work, we challenge this paradigm by returning to mathematical first principles. We propose the \textbf{Clifford Algebra Network (CAN)}, also referred to as \modelname, a vision backbone grounded purely in Geometric Algebra. Instead of engineering separate modules for mixing and memory, we derive a unified interaction mechanism based on the \textbf{Clifford Geometric Product} ($uv = u \cdot v + u \wedge v$). This operation ensures algebraic completeness regarding the Geometric Product by simultaneously capturing feature coherence (via the generalized inner product) and structural variation (via the exterior wedge product). 

Implemented via an efficient sparse rolling mechanism with \textbf{strict linear complexity $\mathcal{O}(N)$}, our model reveals a surprising emergent property: the geometric interaction is so representationally dense that standard Feed-Forward Networks (FFNs) become redundant. Empirically, \modelname{} establishes a new Pareto frontier: our \textbf{Nano} variant achieves \textbf{77.82\%} accuracy on CIFAR-100 with only \textbf{1.4M} parameters, effectively matching the heavy-weight ResNet-18 (11.2M) with \textbf{$8\times$ fewer parameters}, while our \textbf{Lite} variant (2.6M) sets a new SOTA for tiny models at \textbf{79.05\%}. Our results suggest that global understanding can emerge solely from rigorous, algebraically complete local interactions, potentially signaling a shift where \textit{geometry is all you need}. Code is available at \url{https://github.com/ParaMind2025/CAN}.
\end{abstract}

\vspace{2ex}
\section{Introduction}
\vspace{-9ex}
\epigraph{``The two systems [Hamilton's and Grassmann's] are not only consistent with one another, but they are actually parts of a larger whole."}{--- \textup{William Kingdon Clifford}, 1878}
\vspace{1ex}

% ----------------------------------------------------------------------
% 1. INTRODUCTION
% ----------------------------------------------------------------------

\lettrine[lines=2]{T}he design of deep neural networks has long been dominated by a dichotomy: engineering empiricism versus physical mimicry. On one hand, architectures like the Vision Transformer (ViT) \cite{dosovitskiy2020image} rely on the stacking of engineered blocks (Self-Attention + MLP), a pattern now known as ``MetaFormer'' \cite{yu2022metaformer}. On the other hand, recent works attempt to ground deep learning in physical laws, modeling networks as discretizations of diffusion processes or fluid dynamics.

\textbf{Beyond Physical Mimicry: A Mathematical Ansatz.} While drawing inspiration from physics is valuable, we argue that deep learning architecture design need not be constrained by specific physical laws. Instead, we seek freedom in the abstract landscape of mathematics. We propose a governing equation driven not by a specific physical force, but by \textbf{algebraic completeness}. It represents a mathematically principled \textit{Ansatz}: that the evolution of latent representations should be driven by the full spectrum of geometric relationships — both scalar alignment (similarity) and bivector structure (orthogonality) — available in the vector space. To this end, we adopt \textbf{Geometric Algebra (Clifford Algebra)} \cite{hestenes1966space, dorst2007geometric} as the foundational language of our architecture.

\textbf{Emergent Globality from Local Completeness.}
A prevailing dogma in modern vision is the necessity of explicit global context (e.g., Global Self-Attention) to capture long-range dependencies. We challenge this view. We posit that \textbf{global understanding is an emergent property of rigorous local processing}. By utilizing Geometric Algebra to maximize the information extraction from \textbf{local context} — creating a dense ``entanglement'' between a feature and its neighborhood — we implicitly encode global structure without the quadratic cost of global retrieval. 

Building on this philosophy, we introduce \modelname. Unlike standard Transformers that require heavy Feed-Forward Networks (FFNs) to perform channel mixing and non-linear transformation, \modelname{} relies on a \textbf{Dual-Stream Geometric Block} powered by the Clifford Geometric Product. One stream captures high-frequency details, while the other aggregates local context; their interaction is modeled via the \textbf{geometric product} to drive feature updates. This interaction mechanism is so expressive that it allows us to drastically reduce, or even \textbf{completely remove}, the FFN component.

Crucially, to make this geometric paradigm feasible for high-resolution and high-dimensional backbones, we introduce a \textbf{Sparse Rolling Interaction} strategy. Rather than computing the full $D \times D$ outer product matrix (which scales quadratically with channel dimension), we sample the tangent space via cyclic shifts. This ensures the overall complexity remains linear $\mathcal{O}(N)$ in sequence length and $\mathcal{O}(D)$ in channel dimension.

Our contributions are four-fold:

\begin{itemize}
    \item \textbf{Mathematical Unification via Clifford Algebra:} We reframe visual feature interaction through the lens of \textit{Algebraic Completeness}. By proposing the \textbf{Clifford Interaction Ansatz} based on the full Geometric Product ($uv = u \cdot v + u \wedge v$), we restore the missing spatial structure (bivector) alongside standard scalar similarity, unifying feature gating and geometric flow into a single, theoretically rigorous operation.
    
    \item \textbf{Geometric Evolution via Local Context:} We formulate deep representation learning asa continuous \textbf{dynamic evolution governed by a differential equation}. Driven by local geometric context (approximated via Laplacian operators), this ansatz bridges physical diffusion processes with neural representation learning, allowing the network to adaptively model feature coherence and structural variation.
    
    \item \textbf{Native 2D Topological Fidelity:} We challenge the prevailing trend of \textit{image serialization} (flattening in ViTs, scanning in SSMs). \modelname{} operates natively on isotropic 2D feature grids, utilizing sparse rolling interactions that inherently respect spatial adjacency. This design preserves the intrinsic topology of visual data without requiring complex positional encodings or artificial scanning heuristics.
    
    \item \textbf{Paradigm Shift in Efficiency:} We challenge the prevailing meta-architecture by demonstrating that heavy Feed-Forward Networks (FFNs) are redundant when geometric interactions are sufficiently expressive. Our \textit{No-FFN} \modelname{} achieves high-level accuracy with only \textbf{2.6M} parameters, significantly outperforming MobileNetV2, ShuffleNetV2 and ViT-Tiny while establishing a new Pareto frontier for linear-complexity backbones.
\end{itemize}

\footnotetext{
Definition: Algebraic Completeness under Geometric Product.
We use the term \textit{algebraic completeness} to denote the utilization of the \textbf{full} Clifford Geometric Product between two vectors. Standard neural primitives typically utilize only the symmetric scalar component ($\mathbf{u} \cdot \mathbf{v}$), discarding the anti-symmetric bivector component ($\mathbf{u} \wedge \mathbf{v}$). An architecture is considered algebraically complete in this context if it explicitly models both the coherence (scalar) and structural (bivector) terms of the interaction.
}
% ----------------------------------------------------------------------
% 2. RELATED WORK
% ----------------------------------------------------------------------
\section{Related Work}

\textbf{Geometric \& Physics-Inspired Dynamics.}
Recent research has increasingly integrated geometric priors and physical analogies into deep learning architectures. A pivotal example is \textbf{RiemannFormer} \cite{ji2025riemannformerframeworkattentioncurved}, which explicitly models the latent space as a Riemannian manifold, employing parallel transport to rectify attention mechanisms. Building upon this geometric foundation, Di Sipio et al. \cite{disipio2025curvedspacetimetransformerarchitectures} extended the analogy to General Relativity, interpreting attention weights as a metric that induces curvature in a spacetime where layers represent discrete time steps.

While these approaches offer robust theoretical frameworks, they often rely on explicit and computationally intensive manifold operations. \modelname{} shares the dynamic view of treating network depth as temporal evolution but proposes a fundamental paradigm shift in the governing equation. Instead of simulating gravitational curvature or explicitly constructing manifold operators to \textit{support} attention, we internalize geometry directly into the feature interaction via a \textbf{Geometric Algebra ansatz}. By driving state updates through local geometric conflict (wedge product) and alignment (dot product), we move from ``Geometry for Attention'' to ``Geometry as Computation,'' \textbf{achieving linear complexity without sacrificing theoretical depth}.

\noindent \textbf{Clifford Neural Networks.}
The integration of Clifford Algebra into deep learning has gained traction, particularly for tasks requiring geometric symmetry. Existing works \cite{brandstetter2023geometric, brandstetter2023clifford} largely focus on 3D physics simulations and PDE solving, aiming for E(3)-equivariance. However, these architectures typically treat network weights themselves as multivectors, resulting in a combinatorial explosion of parameters (e.g., 16 components for 4D spacetime algebra) and heavy matrix-vector multiplication overhead.
In contrast, \modelname{} repurposes Geometric Algebra for \textit{general-purpose 2D vision}. We do not use Clifford-valued weights; instead, we employ the Geometric Product purely as a lightweight $\mathcal{O}(N)$ \textbf{token interaction mechanism} over real-valued feature maps. This design choice allows us to harness the representational density of bivectors without the computational penalty of full algebraic implementations, making it feasible for high-resolution image recognition.

\noindent \textbf{Efficient Backbones \& Linear Models.}
The pursuit of efficiency has bifurcated into two paths. 
On one hand, CNN-based architectures like \textbf{MobileNetV2} \cite{sandler2018mobilenetv2} and \textbf{ShuffleNetV2} \cite{ma2018shufflenet} optimize local operations via depthwise separability and channel shuffling. While efficient, they lack global context modeling. 
On the other hand, \textbf{ViT-Tiny} \cite{touvron2021training} introduces global attention but suffers from quadratic complexity $\mathcal{O}(N^2)$, limiting scalability.
More recently, linear-complexity models like \textbf{Mamba (SSM)} \cite{gu2023mamba} and its vision counterparts \cite{zhu2024vision,liu2024vmamba}, and \textbf{RWKV} \cite{peng2023rwkv} have emerged, replacing attention with recurrent state-space models.
\modelname{} distinguishes itself from these paradigms. Unlike CNNs, it captures long-range dependencies via rolling interactions; unlike ViTs, it maintains strict linear complexity; and unlike SSMs which rely on recurrent state compression, our model relies on \textbf{algebraic density}. Notably, we achieve superior parameter efficiency by rendering the heavy FFN block redundant — a feat not yet demonstrated in CNN or SSM architectures.

\noindent \textbf{Native 2D Topology vs. Serialization.} 
A fundamental challenge in modern vision is adapting sequence models to 2D images. 
\textbf{ViTs} flatten spatial dimensions, relying on global attention to recover lost topological neighborhoods. 
\textbf{VMamba} \cite{liu2024vmamba} and other SSM-based visual backbones address this by introducing complex scanning trajectories (e.g., cross-scanning) to linearize the image while attempting to preserve 2D causality. While efficient, these scanning heuristics break the continuous manifold structure of the image into discrete 1D paths.

In contrast, \modelname{} operates \textbf{natively} on the 2D feature grid. Our geometric context $\mathcal{C}(H)$ is instantiated via depth-wise convolution, which inherently respects the spatial locality and isotropy of the image manifold without requiring artificial linearization. We posit that maintaining this \textbf{topological fidelity} is a key factor in our model's superior performance-efficiency ratio compared to serialization-based approaches.

% ----------------------------------------------------------------------
% 3. METHODOLOGY
% ----------------------------------------------------------------------
\section{Methodology}

\lettrine[lines=2]{I}n this work, we propose a new family of architectures, the \textbf{Clifford Algebra Networks (CANs)}, which leverage the rich structure of \textbf{Geometric Algebra} to drive \textbf{feature evolution}. While the full algebraic space offers a vast landscape of geometric operations (e.g., multivector convolutions, rotations via rotors), we begin by instantiating the \textbf{first generation} of CANs, which we term \textbf{\modelname{}}. 

\modelname{} is built upon the most fundamental operation in Clifford Algebra: the \textbf{Geometric Product} ($uv = u \cdot v + u \wedge v$). The core hypothesis is that this algebraically complete interaction is sufficient to replace the decoupled mixing stages of standard backbones. In the following part of this section, we present the theoretical foundation of our approach, moving from a general ansatz of feature evolution to the specific instantiation using Clifford Algebra.

\subsection{Preliminaries}

Given an input image $I \in \mathbb{R}^{\mathbb{H} \times \mathbb{W} \times 3}$, we apply a patch embedding layer (typically a convolution with kernel size and stride $P$) to project pixel patches into a latent feature space. 
\textbf{Crucially, unlike standard Vision Transformers that flatten patches into a 1D sequence, we maintain the 2D spatial arrangement} of the tokens throughout the network to facilitate geometric context modeling.

We denote the feature state at a given layer as $X \in \mathbb{R}^{h \times w \times D}$, where:
\begin{itemize}
    \item $h = \mathbb{H}/P$ and $w = \mathbb{W}/P$ represent the feature map resolution.
    \item $D$ denotes the channel (feature) dimension.
\end{itemize}

\noindent \textbf{Isotropic Architecture.}
Unlike hierarchical backbones (e.g., ResNet, Swin) that progressively reduce spatial resolution while increasing channel width, \modelname{} adopts an \textbf{isotropic (columnar) design}.
Upon the initial patch projection, the feature map dimensions $(h \times w \times D)$ remain constant throughout all $L$ layers. This design choice aligns with our theoretical formulation of feature learning as a continuous geometric flow within a fixed phase space, allowing for deep stacking of interaction blocks without spatial information loss. We focus on designing an interaction mechanism that scales linearly with respect to both $N=hw$ and $D$. 

Throughout this paper, unless otherwise specified, we formulate our equations in a pointwise manner. Let $H(t) \in \mathbb{R}^D$ denote the feature vector of a specific token at layer (time) $t$. 
The global feature map is thus a vector field $\mathbf{H} : \Omega \to \mathbb{R}^D$ defined over the spatial grid $\Omega$.
Operations such as the Clifford product are applied pointwise, while the context operator $\mathcal{C}(\cdot)$ (instantiated via convolution) aggregates spatial information from the local neighborhood of the vector field.

\subsection{A General Framework for Feature Evolution}

We model the layer-wise update of visual features as \textbf{a continuous dynamic evolution process}. Specifically, we hypothesize that the evolution of $H$ follows a differential equation governed by the \textbf{interaction between the current state and its geometric context}:

\begin{equation}
    \label{eq:general_ode}
    \frac{\partial H}{\partial t} = \mathcal{F}\Big( H, \mathcal{C}(H) \Big)
\end{equation}

\vspace{0.3em}
Here, $\mathcal{C}(H)$ represents the contextual field derived from $H$, and $\mathcal{F}$ is an interaction function coupling the state and its context. This formulation generalizes Residual Networks, where the residual branch approximates the derivative $\frac{\partial H}{\partial t}$. Our core innovation lies in the specific design of $\mathcal{C}$ and $\mathcal{F}$ to incorporate explicit geometric reasoning.

\subsection{The Clifford Interaction Ansatz}

To fully capture the geometric relationship between the state $H$ and its context $\mathcal{C}$, we define the interaction $\mathcal{F}$ using the \textbf{Clifford Geometric Product}\footnotemark. 

\footnotetext{
    \textbf{Historical Note:} This unification traces back to William Kingdon Clifford (1878), who realized that Hamilton's algebra of rotations (quaternions, relying on inner products) and Grassmann's algebra of extensions (subspaces, relying on outer products) were ``not only consistent with one another, but parts of a larger whole.'' Modern deep learning has inadvertently repeated this historical schism by relying heavily on the dot product (Attention) while neglecting the wedge product. This work aims to restore that algebraic unity.
}

Standard neural network operations (like dot-product attention) typically project interactions onto a scalar field. We argue that this is geometrically lossy. We define the interaction force as a projection of the full geometric product:
\begin{equation}
    \mathcal{F}(H, \mathcal{C}) \coloneqq \mathcal{P}\left( \underbrace{H \cdot \mathcal{C}}_{\text{Coherence}} \oplus \underbrace{H \wedge \mathcal{C}}_{\text{Structure}} \right)
\end{equation}

\vspace{0.3em}
where $\mathcal{P}$ is a learnable projection operator mapping the multi-grade Clifford space (scalars + bivectors) back to the vector space $\mathbb{R}^D$. This formulation unifies two distinct geometric priors:

\begin{enumerate}
    \item \textbf{The Generalized Inner Product ($H \cdot \mathcal{C}$):} This term captures the \textit{alignment} or similarity between the feature and its context. In our implementation, we compute this as an element-wise Hadamard product followed by a linear mixing. It acts as a \textbf{gating mechanism} or a diffusion force, regulating feature magnitude based on coherence.
    
    \item \textbf{The Exterior Product ($H \wedge \mathcal{C}$):} This term constructs a bivector (2-blade) representing the plane spanned by $H$ and $\mathcal{C}$. It resides on the \textbf{Grassmann manifold $\text{Gr}(2, D)$}. This term captures \textit{orthogonality} and structural variation. Physically, it acts as a geometric torque or vorticity, inducing rotations in the feature space to highlight edges and texture boundaries where the local context diverges from the center.
\end{enumerate}

\vspace{0.5em}
\noindent \textbf{Primer on Bivector Bases.}
To demystify the exterior product, consider the input features in a standard basis $\{ \mathbf{e}_1, \dots, \mathbf{e}_D \}$. The wedge product $\mathbf{u} \wedge \mathbf{v}$ yields a \textbf{bivector}, which can be expanded as a linear combination of basis bivectors:

\begin{equation}
    \mathbf{u} \wedge \mathbf{v} = \sum_{1 \le i < j \le D} (u_i v_j - v_i u_j) (\mathbf{e}_i \wedge \mathbf{e}_j)
\end{equation}

Here, each basis element $\mathbf{e}_i \wedge \mathbf{e}_j$ represents a unit oriented plane spanning dimensions $i$ and $j$. While the full bivector space has dimension $\binom{D}{2}$, our architecture does not compute all coefficients. Instead, it selectively computes those corresponding to specific spatial/channel relationships defined by the rolling shifts, ensuring efficiency while capturing essential geometric structures.

\vspace{0.5em}
\noindent \textbf{Remark: Vectorization of the Geometric Product.} 
Strictly speaking, the standard Clifford geometric product yields a \textit{multivector} residing in a graded algebra space (mixing scalars and bivectors). To adapt this algebraic structure to the vector-space formalism of deep neural networks, we interpret the direct sum $\oplus$ as a \textbf{channel-wise concatenation}. 
This operation effectively embeds the heterogeneous multivector components into a unified high-dimensional feature space ($\mathbb{R}^{2D}$). The subsequent projection $\mathcal{P}$ then acts as a learnable \textit{contraction} operator, fusing the coherent (scalar-derived) and structural (bivector-derived) information back into the fundamental latent dimension. This formulation allows us to harness the algebraic completeness of Clifford theory while maintaining compatibility with standard backpropagation dynamics.

% -----------------------------------------------------------
% Section Context Definition
% -----------------------------------------------------------

\subsection{Context Instantiation: Local-Global Duality}
\label{sec:context}

The geometric context $\mathcal{C}(H)$ serves as the reference field driving feature evolution. In our framework, $\mathcal{C}$ is not constrained to a single scale; rather, it can be instantiated to capture either \text{local structural variation} or \textbf{global semantic coherence}, depending on the design constraints.

\begin{itemize}
    \item \textbf{Local Context (Implicit Globality):} For extreme efficiency, $\mathcal{C}$ can be instantiated via local operators (e.g., convolutions). Here, the global receptive field emerges implicitly through the layer-wise propagation of local interactions, akin to a diffusion process.
    
    \item \textbf{Global Context (Explicit Globality):} To maximize performance, $\mathcal{C}$ can be derived from a global summary (e.g., global average, linear attention). This mirrors the global information flow in Transformers but differs fundamentally in complexity. While Transformers rely on quadratic Attention ($\mathcal{O}(N^2)$), our \textbf{Global Clifford Interaction} remains strictly linear ($\mathcal{O}(N)$), interacting the local state $H_i$ with the global mean field $\mathcal{C}_{glo}$.
\end{itemize}

\noindent \textbf{Design Choice.} 
We posit that the \textbf{Global Context} variant is the natural geometric successor to the standard Feed-Forward Network (FFN). While FFNs mix channels blindly, the Global Clifford Interaction aligns local features with the global scene context geometrically. Thus, for our high-performance variants, we employ a superposition of both scales.

\vspace{0.5em}
\noindent \textbf{1. Instantiating the Local Context (Diffusion).}

\vspace{0.5em}
\noindent \textbf{Factorized Linear Laplacian (The Smoothing Step).} 
To capture macroscopic structures with a large receptive field (e.g., $7 \times 7$) while optimizing parameter efficiency, we adopt a \textbf{factorized} strategy. We define the local context field $\mathcal{C}_{loc}$ as a stack of two smaller depth-wise convolutions separated by non-linear activation:

\begin{equation}
    \mathcal{C}_{loc}(H) = \text{Conv}_{3 \times 3} \Big( \text{Conv}_{3 \times 3}(H) \Big)
\end{equation}

This design is \textbf{hardware-aware}: it leverages highly optimized $3 \times 3$ convolution primitives (e.g., cuDNN) to ensure high throughput, avoiding the latency overhead of sparse or irregular operators.

\vspace{0.5em}
\noindent \textbf{Generalized Geometric Prior (The Adaptation Step).}
We further generalize the context definition by introducing a \textit{self-energy suppression factor} $\lambda \in \{0, 1\}$. The final context $\mathcal{C}$ entering the interaction layer is defined as:

\begin{equation}
    \mathcal{C} = \mathcal{C}_{loc}(H) - \lambda \cdot H
\end{equation}

This formulation unifies two distinct geometric modes based on model capacity:
\begin{itemize}
    \item \textbf{Differential Mode ($\lambda=1$):} Here, the context effectively becomes the \textbf{discrete Laplacian} $\mathcal{C} \approx \Delta H$. This formulation acts as a pure \textbf{geometric high-pass filter}, explicitly suppressing the base feature intensity from the identity path. We find this optimal for capacity-constrained models, as it maximizes the signal-to-noise ratio of structural variations.
    \item \textbf{Absolute Mode ($\lambda=0$):} $\mathcal{C} = \mathcal{C}_{loc}$. This acts as an \textbf{energy-preserving flow}. The subsequent inner product ($H \cdot \mathcal{C} \approx \|H\|^2 + H \cdot \Delta H$) retains feature intensity information.
\end{itemize}

\vspace{0.5em}
\noindent \textbf{2. Global-Local Geometric Superposition}

While the local context effectively captures intricate local geometric structures, integrating the global context can further enhance the overall representation. To achieve this without the quadratic cost of self-attention, we introduce a \textbf{Dual-Scale Superposition} mechanism.

We define a global context stream $\mathcal{C}_{glo}$ derived from the global average of the feature field:

\begin{equation}
    \mathcal{C}_{glo} = \text{GlobalAvgPool}(H)
\end{equation}

We then compute the Clifford interaction between the state $H$ and this global context. This operation, which we term \textbf{geometric FFN-Global (gFFN-G)}, serves as a geometrically interpretable alternative to standard MLPs. gFFN-G computes the full geometric product $H \mathcal{C}_{glo}$. The \textbf{Wedge term} ($H \wedge \mathcal{C}_{glo}$) specifically highlights features that deviate from the global background (saliency detection), while the \textbf{Inner term} ($H \cdot \mathcal{C}_{glo}$) reinforces features consistent with the global scene.

\vspace{0.5em}
\noindent \textbf{Unified Superposition Principle.}
Following the physical principle of field superposition, we integrate the local and global geometric forces via a weighted summation. The governing evolution equation is thus generalized as:

\begin{equation}
    \label{eq:superposition}
    \frac{\partial H}{\partial t} = \mathcal{P}_{loc}\Big( H (\mathcal{C}_{loc}) \Big) + \beta \cdot \mathcal{P}_{glo}\Big( H (\mathcal{C}_{glo}) \Big)
\end{equation}

where: $\mathcal{C}_{loc} \approx \Delta H$ represents the local differential context (via factorized convolutions). $\mathcal{C}_{glo} = \text{GlobalAvg}(H)$ represents the global mean field. $\beta \in \{0, 1\}$ serves as a \textbf{structural switch}: $\beta=0$ prioritizes extreme parameter efficiency by relying solely on local geometric diffusion; $\beta=1$ integrates the global interaction mechanism to capture long-range semantic dependencies.

This design allows the network to simultaneously sharpen local boundaries (via local curvature) and highlight salient objects (via global contrast), unifying micro and macro perception in a single, linear-complexity update step.

\vspace{0.5em}
\noindent \textbf{3. Unified Algebraic Mechanism.}

Crucially, despite the distinct physical interpretations of $\mathcal{C}_{loc}$ (local curvature) and $\mathcal{C}_{glo}$ (global mean field), both terms in Eq. (\ref{eq:superposition}) are governed by an identical algebraic mechanism: the Clifford Geometric Product. 

Therefore, in the following sections, we focus our derivation on the general interaction form $\mathcal{F}(H, \mathcal{C}) = \mathcal{P}(H(\mathcal{C}))$. Whether $H$ interacts with its neighbor or the global context, the underlying operation remains the same: a unification of the generalized inner product (coherence) and the exterior product (structure):

\begin{equation}
    \label{eq:evolution_decomp}
    \frac{\partial H}{\partial t} = \mathcal{P}\Big( H (\mathcal{C}(H)) \Big) = \mathcal{P}\Big( \underbrace{\mathcal{D}(H, \mathcal{C})}_{\text{Scalar Component}} \oplus \underbrace{\mathcal{W}(H, \mathcal{C})}_{\text{Bivector Component}} \Big)
\end{equation}

\vspace{0.3em}
\noindent where the operators $\mathcal{D}(\cdot, \cdot)$ and $\mathcal{W}(\cdot, \cdot)$ correspond to the \textbf{Generalized Inner Product} and \textbf{Exterior Product} components of the Clifford geometric product, respectively. 

\vspace{0.3em}
\noindent \textbf{From Full Operators to Sparse Approximations.}
Calculating the full operators $\mathcal{D}$ and $\mathcal{W}$ densely (i.e., all-to-all channel interactions) would incur quadratic complexity $\mathcal{O}(D^2)$. To ensure scalability, we approximate these full operators using a \textbf{Sparse Rolling} strategy. In the following section, we define their efficient discretizations, denoted as $\mathcal{D}_s$ and $\mathcal{W}_s$, parameterized by cyclic shift offsets $s$.

\subsection{Efficient Realization: The Shifted Geometric Product}

Directly computing the full Clifford Geometric Product for all channel pairs requires calculating the complete outer product matrix, which scales quadratically with the channel dimension $\mathcal{O}(D^2)$. To maintain the linear complexity required for high-dimensional vision backbones, we introduce a \textbf{Sparse Rolling Interaction} strategy.

\vspace{0.5em}
\noindent \textbf{The Translation Operator.} 
We generalize the standard geometric product using a channel translation operator $\mathcal{T}_s$, which cyclically shifts the feature vector by an offset $s$. For a set of sparse shift offsets $\mathcal{S}$, we first define the scalar and bivector components using the Hadamard product $\odot$:
\begin{equation}
\begin{aligned}
    \label{eq:rolling_ops}
    \mathcal{D}_s(H, \mathcal{C}) &= \text{SiLU}\Big( H \odot \mathcal{T}_s(\mathcal{C}) \Big) \\
    \mathcal{W}_s(H, \mathcal{C}) &= H \odot \mathcal{T}_s(\mathcal{C}) - \mathcal{T}_s(H) \odot \mathcal{C}
\end{aligned}
\end{equation}

\noindent \textbf{The Geometric Product as a Spectrum.}
Conceptually, the full geometric product $H \mathcal{C}$ encompasses all pairwise interactions between channels $i$ and $j$. This can be visualized as a dense correlation matrix. 
Our efficient realization does not compute this entire matrix; instead, it extracts specific spectral components corresponding to fixed channel offsets. 
Let $[ H \mathcal{C} ]_s$ denote the subset of interaction terms where the channel indices differ by $s$ (i.e., interactions between $h_c$ and $\kappa_{c+s}$). We define this slice as:

\begin{equation}
    [ H \mathcal{C} ]_s \coloneqq \underbrace{\mathcal{D}_s(H, \mathcal{C})}_{\text{Scalar Component}} \oplus \underbrace{\mathcal{W}_s(H, \mathcal{C})}_{\text{Bivector Component}}
\end{equation}

\vspace{0.3em}
By implementing this extraction via the rolling operator $\mathcal{T}_s$, we recover these specific diagonals of the full geometric product with linear complexity.

%%% Figure
%%%
\begin{figure}[!ht]
%\hspace{0in}
\subfigure{
\begin{minipage}[b]{1.0\linewidth}
\centering
\includegraphics[width=0.8\linewidth]{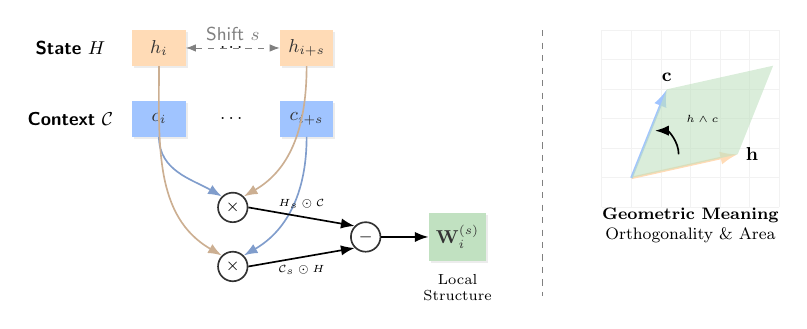} 
\vspace{-0.5em}
\end{minipage}
}
\caption{
Visualizing the Efficient Wedge Product.
    We operationalize the Geometric Product via a linear-complexity rolling mechanism. The Wedge Product branch captures structural variation. Instead of computing a full quadratic tensor, we cyclically shift the feature streams ($\mathcal{T}_s$) to compute cross-term differences. Geometrically, this constructs a \textbf{bivector} (green plane), representing the oriented area and orthogonality between the state $\mathbf{h}$ and context $\mathbf{c}$.}
\label{fig:wedge_viz}
\end{figure}

\noindent \textbf{Implementation Details.} 
In terms of explicit tensor indices, let $H_{i,c}$ and $\mathcal{C}_{i,c}$ denote the feature and context values at spatial position $i$ and channel $c$. The operator formulations above translate to the following element-wise computations:

\begin{equation}
\begin{aligned}
    \label{eq:shifted_indices}
    \text{Dot}^{(s)}_{i,c} &= \text{SiLU}\Big( H_{i,c} \cdot \mathcal{C}_{i, (c+s)\%D} \Big) \\[0.5em]
    \text{Wedge}^{(s)}_{i,c} &= H_{i,c} \cdot \mathcal{C}_{i, (c+s)\%D} - \mathcal{C}_{i,c} \cdot H_{i, (c+s)\%D}
\end{aligned}    
\end{equation}

\vspace{0.3em}
We visually illustrate the mechanism of Wedge Product in Figure \ref{fig:wedge_viz}. By cyclically shifting the feature streams, we effectively pair the current state with channel-wise distant context without incurring random memory access costs. 

\vspace{0.3em}
\noindent \textbf{Interpretation.}
The \textbf{Inner term} (Eq. \ref{eq:shifted_indices} top) corresponds to a \textit{generalized inner product} $\mathbf{u}^\top \mathbf{P}_s \mathbf{v}$, where $\mathbf{P}_s$ is the cyclic permutation matrix corresponding to shift $s$. Enhanced with the SiLU activation, it acts as a non-linear coherence gate, filtering features based on their alignment with the shifted context. 
The \textbf{Wedge term} (Eq. \ref{eq:shifted_indices} bottom) explicitly computes the \textbf{Bivector Coefficients} for the plane spanned by basis vectors $\mathbf{e}_c$ and $\mathbf{e}_{c+s}$. Geometrically, this term represents the oriented area of the parallelogram spanned by the feature vectors. This area is maximized when features are orthogonal, allowing the network to explicitly detect structural variations and edges. The Wedge term explicitly computes the bivector coefficients ($h_i c_{i+s} - c_i h_{i+s}$), strictly adhering to the anti-symmetric definition of the exterior product to capture geometric orientation.

\vspace{0.5em}
\noindent \textbf{Remark on Basis Orientation.} 
It is worth noting that due to the cyclic boundary conditions of the roll operator $\mathcal{T}_s$, the channel indices pairs $(c, (c+s)\%D)$ may wrap around. Since the exterior product is anti-symmetric ($\mathbf{e}_i \wedge \mathbf{e}_j = -\mathbf{e}_j \wedge \mathbf{e}_i$), wrapped terms effectively correspond to a negative basis orientation compared to a canonical ordered basis (where $i<j$). 
However, explicit sorting to enforce canonical ordering is computationally wasteful and breaks vectorization. In our design, these sign flips are absorbed by the subsequent learnable linear projection $\mathcal{P}$. The network automatically learns to interpret the signed bivector coefficients, allowing us to maintain a purely vectorized, hardware-efficient implementation without branching logic.

By approximating the dense metric tensor with this sparse set of shifted interactions, we reduce the computational complexity from quadratic to linear $\mathcal{O}(N \cdot D \cdot |\mathcal{S}|)$, making \modelname{} highly scalable while retaining rigorous geometric expressivity.

\vspace{-0.4em}
\subsubsection*{Structural Inductive Bias vs. Learned Topology}

Mathematically, the shifted inner product term $\mathbf{u} \cdot (\mathcal{T}_s \mathbf{v})$ can be formalized as a \textbf{bilinear form} $\mathbf{u}^\top \mathbf{P}_s \mathbf{v}$. Unlike the standard inner product (where $\mathbf{P}_s=\mathbf{I}$), this formulation generalizes the notion of similarity. Here, $\mathbf{P}_s$ acts as an adjacency operator defining a specific \textbf{topological connectivity} between channels. This allows the network to measure coherence not just between identical features, but across semantically related yet spatially distant feature channels encoded at specific intervals $s$.

One might ask: why predefine the shifts $\mathcal{S}$ instead of learning a full interaction matrix $\mathbf{A} \in \mathbb{R}^{D \times D}$ to parameterize the bilinear form $\mathbf{u}^\top \mathbf{A} \mathbf{v}$? Learning a full matrix would incur a quadratic parameter cost $\mathcal{O}(D^2)$ and computational complexity, essentially reintroducing the heavy channel mixing of standard FFNs/Attention which we aim to eliminate. Instead, we impose a strong \textbf{structural inductive bias}: we assume that the latent feature space possesses a ring-like topology, where semantic dependencies can be efficiently captured via a sparse set of multi-scale interactions.

By restricting the connectivity to a fixed set of exponential shifts (e.g., $s \in \{1, 2, 4, \dots\}$), we enforce a \textbf{structured sparsity} constraint. This design allows the network to propagate information globally across the channel dimension with logarithmic path length, achieving the expressivity of dense mixing with only linear $\mathcal{O}(D)$ complexity.

\vspace{0.5em}
\noindent \textbf{Matrix Interpretation: Circulant Sparsity.}
The rolling interaction can be rigorously interpreted through the lens of structured matrices. 
Conceptually, the full channel-wise interaction between state $H$ and context $\mathcal{C}$ forms a dense correlation matrix $\mathbf{M} \in \mathbb{R}^{D \times D}$, where $M_{i,j}$ represents the interaction between channel $i$ and $j$.

Our cyclic shift operation ($\mathcal{T}_s$) effectively samples specific diagonals from this matrix. 
If we considered all possible shifts $s \in \{0, \dots, D-1\}$, the operation would be equivalent to a multiplication by a Circulant Matrix (or Toeplitz matrix without wrapping). 
By restricting $s$ to a sparse set $\mathcal{S}$ (e.g., logarithmic steps), we enforce a \textbf{Structured Circulant Sparsity}. This implies that we approximate the full dense geometry by only retaining the most significant frequency bands in the channel interaction topology, reducing complexity from $\mathcal{O}(D^2)$ to $\mathcal{O}(|\mathcal{S}|D)$.

\subsection{Gated Geometric Residual (GGR)}

To translate the continuous geometric evolution (Eq. \ref{eq:general_ode}) into a computational graph, we apply a first-order \textbf{Euler discretization}. By treating the network depth as discrete time steps $t=l$, the update rule naturally forms a residual connection:

\begin{equation}
    \label{eq:euler_step}
    H_{l} = H_{l-1} + \Delta t \cdot \mathcal{F}(H_{l-1}, \mathcal{C}_{l-1})
\end{equation}

\vspace{0.3em}
Here, the step size $\Delta t$ corresponds to a learnable scaling factor (implemented via LayerScale $\gamma$). While this formulation recovers the standard ResNet paradigm, we find that directly adding the raw geometric interaction to the identity path is suboptimal due to noise in the semantic stream.

\vspace{0.5em}
\noindent \textbf{Refined Update Rule.} 
To address this, we propose the \textbf{Gated Geometric Residual (GGR)}. We modify the discretization step to include a non-linear pre-filtering of the state and a gated injection of the geometric force. Formally, let $H_{geo}$ denote the output of the Clifford interaction layer. The update rule becomes:

\begin{equation}
    H_{l} = H_{l-1} + \gamma \odot \Big( \text{SiLU}(H_{l-1}) + \text{Gate}(H_{l-1}, H_{geo}) \odot H_{geo} \Big)
\end{equation}

\vspace{0.3em}
This design acts as a \textbf{stabilized numerical solver}: the $\text{SiLU}(H_{l-1})$ term conditions the current state by suppressing background noise (negative values) before integrating the geometric updates, ensuring that the evolution is driven by salient features. 

\subsection{The gFFN Family: A Generalized Interaction Module}
\label{sec:geoffn}

Standard architectures typically decouple feature processing into spatial mixing (e.g., Attention, Conv) and channel mixing (FFN/MLP). We challenge this dichotomy by proposing the \textbf{geometric Feed-Forward Network (gFFN)}. 

gFFN acts as a \textbf{Spatial-Channel Entangler}. By leveraging the Clifford geometric product, it simultaneously performs channel mixing (via projection) and spatial modulation (via context interaction). We unify our architectural variants under this framework by defining three modes of geometric context $\mathcal{C}$:

\begin{itemize}
    \item \textbf{gFFN-L (Local):} $\mathcal{C} = \Delta H$. Utilizes local difference operators to capture high-frequency texture and edges. This powers our \textbf{Nano} variant, maximizing efficiency.
    
    \item \textbf{gFFN-G (Global):} $\mathcal{C} = \text{GlobalAvg}(H)$. Utilizes the global mean field to enforce semantic coherence. 
    
    \item \textbf{gFFN-H (Hybrid):} $\mathcal{C} = \Delta H + \beta \cdot \text{GlobalAvg}(H)$. Combines micro and macro contexts via field superposition. This powers our \modelname{} variants, achieving SOTA performance by integrating multi-scale geometric priors.
\end{itemize}

This modular design allows \modelname{} — and \textbf{potentially other architectures} — to flexibly trade off between local precision and global awareness based on computational constraints.

\vspace{0.36em}
\subsection{Algorithmic Summary}

To provide a holistic view of this paradigm shift, we present the complete forward pass of the \textbf{\modelname{} Block} in Algorithm \ref{alg:clifford_block}. 
This algorithm explicitly details the \textit{No-FFN variant}, illustrating how the distinct geometric components — Dual-Stream generation, Sparse Rolling Interaction, and Gated Geometric Residual — orchestrate to evolve features. 
Notably, the workflow terminates immediately after the geometric fusion (Line 34), bypassing the computational and parameter overhead of the traditional MLP block used in MetaFormer architectures.

\begin{algorithm}[t]
\caption{Pseudo-code for CliffordNet Block (No-gFFN-G Variant)}
\label{alg:clifford_block}
\small
\renewcommand{\algorithmicensure}{\textbf{Output:}}
\renewcommand{\algorithmicrequire}{\textbf{Input:}}

\begin{minipage}[t]{0.49\textwidth}
\begin{algorithmic}[1] 

\REQUIRE Feature map $\mathbf{X}_{l-1}$: \textcolor{shapecolor}{$(\mathtt{B}, \mathtt{H}, \mathtt{W}, \mathtt{C})$}
\ENSURE Updated map $\mathbf{X}_{l}$ : \textcolor{shapecolor}{$(\mathtt{B}, \mathtt{H}, \mathtt{W}, \mathtt{C})$}

\STATE \textcolor{lgreen}{\text{/* 1. Input Norm */}}
\STATE $\mathbf{X}_{ln}$ $\leftarrow$ $\mathbf{LayerNorm}(\mathbf{X}_{l-1})$

\STATE \textcolor{lgreen}{\text{/* 2. Dual-Stream Generation */}}
\STATE $\mathbf{Z}_{det}$ $\leftarrow$ $\mathbf{Linear}_{det}(\mathbf{X}_{ln})$
\STATE $\mathbf{Z}_{ctx}$ $\leftarrow$ $\mathbf{SiLU}(\mathbf{BN}(\mathbf{DWConv}(\mathbf{X}_{ln})))$

\IF{$\text{ctx\_mode} = '\!\text{diff}\,'$}
    \STATE $\mathbf{Z}_{ctx} \leftarrow \mathbf{Z}_{ctx}-\mathbf{Z}_{det}$
\ENDIF

\STATE \textcolor{lgreen}{\text{/* 3. Efficient Interaction (Rolling) */}}
\STATE $\mathbf{F}_{list} \leftarrow []$
\FOR{$s$ in $\mathcal{S}$ \text{ (e.g., \{1, 2, 4, ...\})}}
    \STATE $\mathbf{Z}_{d}^{(s)} \leftarrow \mathbf{Roll}(\mathbf{Z}_{det}, s)$
    \STATE $\mathbf{Z}_{c}^{(s)} \leftarrow \mathbf{Roll}(\mathbf{Z}_{ctx}, s)$
    \STATE $\mathbf{W}^{(s)} \leftarrow \mathbf{Z}_{det} \odot \mathbf{Z}_{c}^{(s)} - \mathbf{Z}_{ctx} \odot \mathbf{Z}_{d}^{(s)}$
    \STATE $\mathbf{D}^{(s)} \leftarrow \mathbf{SiLU}(\mathbf{Z}_{det} \odot \mathbf{Z}_{c}^{(s)})$
    \STATE \textcolor{lgreen}{\text{/* Interaction Mode */}}
    \IF{$\text{cli\_mode} = '\!\text{inner}\,'$}
        \STATE $\mathbf{F}_{list}.\mathbf{append}(\mathbf{D}^{(s)})$
    \ELSIF{$\text{cli\_mode} = '\!\text{wedge}\,'$}
        \STATE $\mathbf{F}_{list}.\mathbf{append}(\mathbf{W}^{(s)})$
    \ELSIF{$\text{cli\_mode} = '\!\text{full}\,'$}
        \STATE $\mathbf{F}_{list}.\mathbf{append}(\mathbf{W}^{(s)}, \mathbf{D}^{(s)})$
    \ENDIF
\ENDFOR
% =================
\end{algorithmic}
\end{minipage}
\hfill %
\begin{minipage}[t]{0.49\textwidth}
\begin{algorithmic}[1]
\setcounter{ALC@line}{24} 

\vspace{1.7em} 
\STATE \textcolor{lgreen}{\text{/* Concat \& Project back to $\mathbb{R}^C$ */}}
\IF{$\text{cli\_mode} = '\!\text{full}\,'$}
    \STATE $\mathbf{G}_{raw}$ : \textcolor{shapecolor}{$(\dots, \mathtt{2|\mathcal{S}|C})$} $\leftarrow$ $\mathbf{Cat}(\mathbf{F}_{list})$
\ELSIF{$\text{cli\_mode} \;\text{in} \;\{'\!\text{inner}\,','\!\text{wedge}\,'\}$}      
    \STATE $\mathbf{G}_{raw}$ : \textcolor{shapecolor}{$(\dots, \mathtt{|\mathcal{S}|C})$} $\leftarrow$ $\mathbf{Cat}(\mathbf{F}_{list})$
\ENDIF
\STATE $\mathbf{G}_{feat}$ : \textcolor{shapecolor}{$(\dots, \mathtt{C})$} $\leftarrow$ $\mathbf{Linear}_{proj}(\mathbf{G}_{raw})$

\STATE \textcolor{lgreen}{\text{/* 4. Gated Geometric Residual */}}
\STATE $\mathbf{M}$ $\leftarrow$ $\mathbf{Cat}([\mathbf{X}_{ln}, \mathbf{G}_{feat}])$
\STATE $\mathbf{\alpha}$ $\leftarrow$ $\mathbf{Sigmoid}(\mathbf{Linear}_{gate}(\mathbf{M}))$
\STATE \textcolor{lgreen}{\text{/* Non-linear modulation */}}
\STATE $\mathbf{H}_{mix}$ $\leftarrow$ $\mathbf{SiLU}(\mathbf{X}_{ln}) + \mathbf{\alpha} \odot \mathbf{G}_{feat}$

\STATE \textcolor{lgreen}{\text{/* 5. Output Update */}}
\STATE \textcolor{lgreen}{\text{/* Note: No FFN applied */}}
\STATE $\mathbf{X}_{l}$ $\leftarrow$ $\mathbf{X}_{l-1} + \mathbf{Drop}(\mathbf{\gamma} \odot \mathbf{H}_{mix})$

\STATE Return: $\mathbf{X}_{l}$ 
\STATE \ 
% =================
\end{algorithmic}
\end{minipage}

\end{algorithm}

% ----------------------------------------------------------------------
% 4. EXPERIMENTS
% ----------------------------------------------------------------------
\section{Experiments}

\lettrine[lines=2]{T}o empirically validate the efficacy of our proposed \textbf{Clifford Interaction Ansatz}, we conduct extensive evaluations on the CIFAR-100 benchmark. While smaller in scale than ImageNet-1K, CIFAR-100 serves as a rigorous litmus test for architectural efficiency and generalization, particularly for lightweight models, due to its high class diversity and data sparsity (only 500 images per class). 

We position these experiments as a foundational \textit{proof-of-concept}: confirming that geometric completeness can yield SOTA performance in resource-constrained regimes, paving the way for future scaling to larger datasets. To ensure a strictly fair comparison with established baselines, we adhere to a rigorous training recipe: 200 epochs, AdamW optimizer, cosine annealing, and regularization (AutoAugment, Random Erasing). For \modelname{}, we additionally employ DropPath \cite{huang2016deep} to regulate the training of our geometric interaction layers. In contrast, for classical CNN baselines (e.g., ResNet-18, MobileNetV2), we exclude DropPath to adhere to their canonical training protocols, where regularization is primarily handled by weight decay and data augmentation. By standardizing the optimizer (AdamW) and augmentation pipeline across all models, we ensure a rigorously controlled comparison, evaluating architectural capability under the same modern data regime.

We employed a \textbf{patch\_size of 2} to achieve a favorable trade-off between efficiency and accuracy. While the results presented herein rely on the \textbf{Differential Mode} due to its demonstrated superiority on CIFAR-100, we acknowledge that different data distributions may benefit from distinct geometric priors. Therefore, we preserve both the Absolute and Differential modes in our framework, allowing for flexible adaptation to other datasets.

\begin{table}[t]
    \centering
    \caption{\textbf{Main Results on CIFAR-100.} We compare \modelname{} against established efficient backbones under an identical training recipe. \textbf{Note on Speed:} Baseline CNNs benefit from mature cuDNN kernel optimizations. Despite relying on unoptimized PyTorch primitives (e.g., rolling), \textbf{\modelname{}-Nano} remains highly competitive in training time while delivering significantly superior accuracy with fewer parameters.}
    \label{tab:main_results}
    \setlength{\tabcolsep}{8pt}
    \begin{tabular}{l c c c c c}
    \toprule
    \textbf{Model} & \textbf{Params} & \textbf{FFN?} & \textbf{Shifts} & \textbf{Time (min)} & \textbf{Top-1 Acc} \\
    \midrule
    \multicolumn{6}{l}{\textit{\textbf{Nano Scale}}} \\
    ShuffleNetV2 1.0$\times$ & 1.4M & Yes & - & 66.2 & 74.60\% \\
    \textbf{\modelname{}-Nano} & 1.4M & No & 2 & 79.9 & \textbf{77.82\%} \\
    \midrule
    \multicolumn{6}{l}{\textit{\textbf{Tiny Scale}}} \\
    MobileNetV2 & 2.3M & Yes & - & 64.5 & 70.90\% \\
    ViT-Tiny & 2.7M & Yes & - & 149.9 & 65.87\% \\
    \textbf{\modelname{}-Lite} & 2.6M & No & 5 & 125.0 & \textbf{79.05\%} \\
    \bottomrule
    \end{tabular}
\end{table}

\subsection{Main Results and Efficiency Analysis}

Table \ref{tab:main_results} presents the comparison with state-of-the-art efficient backbones. Our \textbf{\modelname{}-Lite} achieves \textbf{79.05\%}, establishing a new SOTA for models under 3M parameters, surpassing ResNet-18 (11.2M) by a clear margin.

\vspace{0.5em}
\noindent \textbf{The Cost of Geometry vs. Depth.}
We observe an interesting computational trade-off. Transitioning from \textbf{Nano} (2 shifts) to \textbf{Lite} (5 shifts) improves accuracy by $\sim$1.2\% but increases training time by $\sim$56\%. This indicates that the current bottleneck lies in the \textbf{geometric sampling density} (number of shifts). However, with CUDA acceleration applied to the Clifford Interaction, the training times for both the Nano and Lite variants become comparable (57.64 mins vs. 58.27 mins).

\vspace{0.3em}
\noindent \textbf{Baselines and Architectural Paradigms.} 
We benchmark \modelname{} against three representative efficient architectures. 
\textbf{MobileNetV2} \cite{sandler2018mobilenetv2} relies on inverted residuals and depth-wise convolutions. 
\textbf{ViT-Tiny} \cite{touvron2021training} represents the isotropic Transformer paradigm, which we evaluate under a constrained parameter budget (Ratio=1). 
Most notably, we include \textbf{ShuffleNetV2} \cite{ma2018shufflenet}, which exemplifies the pinnacle of \textit{structural engineering} in CNN design. To mitigate the information blockage caused by group convolutions, ShuffleNetV2 introduces a handcrafted ``Channel Shuffle'' operation, enabling cross-group information flow while optimizing Memory Access Cost (MAC). 

\begin{table}[h]
    \centering
    \caption{\textbf{Comparison with Efficient Backbones on CIFAR-100.} \modelname{} outperforms strong baselines across all parameter scales.}
    \label{tab:comparison}
    \begin{tabular}{l c c c}
    \toprule
    \textbf{Model} & \textbf{Params} & \textbf{MLP Ratio} & \textbf{Top-1 Acc} \\
    \midrule
    \multicolumn{4}{l}{\textit{1.5M Scale (Nano)}} \\
    ShuffleNetV2 1.0$\times$ & 1.4M & - & 74.60\% \\
    \textbf{\modelname{}-Nano} & \textbf{1.4M} & \textbf{0.0} & \textbf{77.82\%} \\
    \midrule
    \multicolumn{4}{l}{\textit{2.5M Scale (Tiny)}} \\
    MobileNetV2 & 2.3M & - & 70.90\% \\
    ShuffleNetV2 1.5$\times$ & 2.6M & - & 75.95\% \\
    ViT-Tiny & 2.7M & 1.0 & 65.87\% \\
    \textbf{\modelname{}-Lite} & \textbf{2.6M} & \textbf{0.0} & \textbf{79.05\%}  \\
    \midrule
    \multicolumn{4}{l}{$>$\textit{3.0M Scale (Small)}} \\
    ResNet-18 & 11.2M & - & 76.75\% \\
    ResNet-50 & 23.7M & - & 79.14\% \\
    DenseNet-121 & 7.0M & - & 80.20\%\\
    \textbf{\modelname{}-32}(3 shifts,\text{cli}\_{mode}='full') & \textbf{4.8M} & \textbf{0.0} & \textbf{81.42\%} \\
    \textbf{\modelname{}-64}(5 shifts,\text{cli}\_{mode}='inner') & \textbf{8.6M} & \textbf{0.0} & \textbf{82.46\%} \\
    \bottomrule
    \end{tabular}
\end{table}

\vspace{0.25em}
\noindent \textbf{Analysis.} 
As shown in Table \ref{tab:comparison}, while ShuffleNetV2 achieves a strong accuracy of 74.60\% through its ingenious engineering heuristics, our \textbf{\modelname{}-Nano} surpasses it by \textbf{4.3\%} with comparable parameters. This result suggests that the \textit{implicit} channel mixing induced by our mathematically grounded Clifford interaction (via the shifted geometric product) is more effective than the \textit{explicit}, handcrafted channel shuffling used in traditional efficient CNNs. Our small models  match or outperform the performance of ResNet-18 and MobileNetV2 by a large margin, validating our hypothesis that geometric interactions can replace generic MLPs. To prove that our efficiency stems from the architecture and not the dataset, we trained a standard ViT-Tiny with `mlp\_ratio=1.0`. The model \textbf{collapsed} to 65.87\% accuracy (Table \ref{tab:comparison}), whereas \textbf{\modelname{}-Lite} maintained 79+\%. This confirms that standard attention lacks the geometric density to function without heavy FFNs.

\noindent \textbf{Scalability and Comparison with Heavyweights.}
As shown in Table \ref{tab:comparison}, \modelname{} demonstrates exceptional parameter efficiency. The \textbf{CliffordNet-Lite} (2.6M) outperforms ResNet-18 (11.2M) by \textbf{+2.3\%} while using $\sim$4$\times$ fewer parameters. Scaling the depth yields significant gains: \textbf{CliffordNet-32} (4.8M) surpasses both ResNet-50 and DenseNet-121. Our deeper variant, \textbf{CliffordNet-64} (8.6M), achieves \textbf{82.46\%}, confirming that the geometric evolution mechanism scales robustly to high-performance regimes entirely \textit{without} FFNs.

All the models here are \textbf{trained from scratch}, without using any pre-trained weights or transfer learning techniques, and without using warm-up training epochs.

\vspace{0.5em}
\noindent \textbf{Discussion.} While Vision Transformers are celebrated for their global receptive field, we argue that this "global" capability comes at the cost of \textbf{interaction poverty}. The standard dot-product attention compresses the complex, high-dimensional relationship between two tokens into a single scalar ($ \mathbf{q} \cdot \mathbf{k} $). This operation is fundamentally lossy: it discards the directional and structural information (the bivector component) inherent in the feature space.
Consequently, Transformers must rely on heavy, parameter-inefficient FFNs to reconstruct and process these lost feature dimensions.

In contrast, CliffordNet takes the opposite approach. We prioritize local algebraic completeness over global scalar matching. By operationalizing the full Geometric Product (and recovering the channel topology via shifted generalized inner products), we extract a dense, information-rich representation from local neighborhoods. Our results show that a network built on \textbf{rich local interactions} can outperform one built on \textbf{sparse global attention}, challenging the prevailing dogma that global is always better.

% ----------------------------------------------------------------------
% 5. abs
% ----------------------------------------------------------------------
\subsubsection*{Ablation Studies}
\vspace{-0.3em}

\begin{table*}[t]
    \centering
    \caption{\textbf{Ablation on Interaction Density and gFFN Necessity.} We compare variants of \modelname{} with different geometric shifts, context modes and gFFNs.}
    \label{tab:ablation_main}
    \vspace{2.0mm}
    \begin{tabular}{l c c c c}
    \toprule
    \textbf{Model Variant}  & \textbf{Shifts} & \textbf{Params} & \textbf{Top-1 Acc}  \\
    \midrule
    \textit{Baseline (MobileNetV2)} & - & 2.3M & 70.90\%  \\
    \midrule
    \textbf{\modelname{}-Nano(abs)} \ & 2 ([1,2]) & 1.43M & 76.41\%  \\
    \textbf{\modelname{}-Lite(abs)} \ & 5 ([1..16]) & 2.61M & 77.63\%  \\
    \midrule
    \textbf{\modelname{}-Nano(diff)}  & 2 ([1,2]) & 1.43M & 77.82\%  \\
    \textbf{\modelname{}-Lite(diff)} \ & 5 ([1..16]) & 2.61M & 79.05\%  \\
    \midrule
    \textbf{\modelname{}-Nano(diff)+gFFN-G} & 2 ([1,2]) & 2.22M & 78.22\%  \\
    \textbf{\modelname{}-Lite(diff)+gFFN-G}  & 5 ([1..16]) & 3.40M & 79.57\%  \\   
    \bottomrule
    \end{tabular}
\end{table*}

\noindent \textbf{Impact of Context Mode and Interaction Density.} To disentangle the contributions of geometric interaction density (number of shifts) and interaction modes, we analyze the variants of \modelname{}.
Table \ref{tab:ablation_main} reveals two key insights. First, the \textbf{Differential Mode} (pure $\Delta H$) consistently outperforms the Absolute Mode by $\sim$1.4\%, confirming that representation learning benefits from filtering out static self-energy. Second, increasing the geometric sampling density (Shifts $2 \to 5$) and integrating the Global Geometric FFN (+gFFN-G) yields further gains. Notably, even our minimal \textbf{Nano(diff)} variant achieves \textbf{77.82\%}, surpassing the MobileNetV2 baseline by nearly \textbf{7\%} with 38\% fewer parameters.

\begin{table}[h]
    \centering
    \caption{\textbf{Component Analysis: Scalar Energy vs. Bivector Structure.} We disentangle the Clifford Interaction into Inner-only and Wedge-only variants (both ~1.63M params). 
    The \textbf{Inner} product captures feature magnitude (energy) via diagonal terms ($u_i u_i$), while the \textbf{Wedge} product is strictly anti-symmetric ($u_i \wedge u_i = 0$), relying solely on structural relations. Remarkably, the Wedge variant rivals the Inner variant despite lacking explicit energy information, and their combination (CliffordNet) yields the best performance.}
    \label{tab:component_analysis}
    \setlength{\tabcolsep}{4pt} % 
    \vspace{2.0mm}
    \begin{tabular}{l c c c c}
    \toprule
    \textbf{Interaction} & \textbf{Algebraic} & \textbf{Diagonal} & \textbf{Feature} & \textbf{Top-1} \\
    \textbf{Variant} & \textbf{Type} & \textbf{(Energy)} & \textbf{Focus} & \textbf{Acc} \\
    \midrule
    \textbf{Inner-Only} & Symmetric & \textcolor{green}{\cmark} (Has Energy) & Coherence & 78.17\% \\
    \textbf{Wedge-Only} & Anti-Sym & \textcolor{red}{\xmark} (No Energy) & Structure & 77.76\% \\
    \midrule
    \textbf{CliffordNet} & \textbf{Geometric} & \textcolor{green}{\cmark} & \textbf{Complete} & \textbf{79.05\%} \\
    \bottomrule
    \end{tabular}
\end{table}

\vspace{0.65em}
\noindent \textbf{Inner Product vs. Exterior Product.}
To disentangle the distinct contributions of scalar and bivector components, we compare \textbf{Inner-Only} and \textbf{Wedge-Only} variants under identical settings ($\mathcal{S}=\{1, 2, 4, 8, 16\}$). All reported results here utilize the Differential Mode.

Table \ref{tab:component_analysis} shows that Inner-Only (\textbf{78.17\%}) slightly outperforms Wedge-Only (\textbf{77.76\%}). However, this comparison reveals a profound geometric insight regarding \textbf{Diagonal Energy}:

\begin{enumerate}
    \item \textbf{Inner Product:} The generalized inner product includes near-diagonal correlations. Crucially, it implicitly encodes feature \textit{magnitude} or \textit{energy} (akin to $x_i^2$), which is fundamental for activation-based deep learning.
    \item \textbf{Wedge Product:} The exterior product is strictly anti-symmetric ($x \wedge x = 0$). It inherently discards all self-magnitude (energy) information and relies solely on cross-channel geometric relations.
\end{enumerate}

\noindent \textbf{Remarkable Robustness of Bivectors.}
Despite operating without explicit energy information, the Wedge-Only variant achieves performance nearly on par with the Inner-Only baseline. This suggests that the \textbf{structural topology} (captured by bivectors) is almost as discriminative as the feature intensity itself. The combination of the two (CliffordNet) thus provides the complete picture: Energy (Inner) + Structure (Wedge).

\vspace{0.3em}
\noindent \textbf{Data-Driven Geometric Selection.}
By providing the full spectrum of geometric interactions (both scalar coherence and bivector structure), \modelname{} avoids hard-coding inductive biases regarding which feature type is more discriminative. 
Instead, the learnable projection layer $\mathcal{P}$ acts as a \textbf{soft selector}, dynamically weighting the contribution of the inner and exterior products based on the data distribution. 
For texture-heavy tasks, the network may prioritize the wedge product; for shape-biased tasks, it may favor the inner product. This \textbf{algebraic completeness} ensures that the optimal geometric representation is always within the hypothesis space of the model, maximizing generalization across diverse visual domains.
For resource-constrained deployment, structural pruning could be applied to selectively remove either the scalar or bivector pathways if the target task relies predominantly on one geometric mode.

% ----------------------------------------------------------------------
% 5. DISCUSSION
% ----------------------------------------------------------------------
\section{Discussion}
\label{sec:discussion}

\lettrine[lines=2]{I}n this section, we analyze the theoretical implications of our empirical findings, interpreting why a pure geometric framework can outperform established architectural paradigms, and positioning \modelname{} within the broader landscape of geometric deep learning.

\subsection{Why does the ``No-FFN'' Paradigm Work?}
The most counter-intuitive finding of this work is that \modelname{} achieves a high performance (77.63\% on CIFAR-100) even without Feed-Forward Networks (FFNs), a component previously considered indispensable in MetaFormer architectures for channel mixing and non-linearity. We attribute this success to the \textbf{Algebraic Density} of the Clifford Interaction:

\begin{itemize}
    \item \textbf{Internalized Non-linearity:} In standard ViTs, the Attention mechanism is primarily a linear aggregation (post-softmax), necessitating a heavy FFN ($\text{ratio}=4$) to provide feature transformation capacity. In contrast, our Clifford Geometric Product inherently involves multiplicative second-order terms (the bivector $u \wedge v$ and scalar $u \odot v$). When combined with the non-linear GGR gating, the interaction layer itself acts as a powerful function approximator. The geometric interaction effectively ``internalizes'' the complexity typically delegated to external MLPs.
    
    \item \textbf{Structured vs. Brute-force Mixing:} An FFN performs dense, agnostic channel mixing via large matrix multiplications. Our Shifted Geometric Product performs \textit{structured} mixing based on the ring topology of the feature space. Our results suggest that this geometrically constrained mixing is more sample-efficient than the brute-force connectivity of large MLPs, allowing the network to perform high-level reasoning with significantly fewer parameters.
\end{itemize}

\subsection{Theoretical View: A Geometric Reaction-Diffusion System}

While the Laplacian term $\mathcal{C} \coloneqq \Delta H$ suggests a connection to heat diffusion (smoothing), the presence of the geometric product introduces a more complex dynamic, closely resembling a \textbf{Geometric Reaction-Diffusion System}.

In mathematical biology, \textbf{Turing patterns} arise from the interplay between \textit{diffusion} (spreading) and \textit{reaction} (local non-linear interaction). \modelname{} can be interpreted as a high-dimensional generalization of this morphogenetic process:
\begin{itemize}
    \item \textbf{Anisotropic Diffusion ($H \cdot \mathcal{C}$):} The scalar term, implemented as a generalized inner product, acts as a coherence-based gating mechanism. It functions effectively as \textit{anisotropic diffusion}, smoothing out noise in homogeneous regions where features align with their context, thereby stabilizing the learning dynamics.
    
    \item \textbf{Geometric Reaction ($H \wedge \mathcal{C}$):} The bivector term introduces a rotational ``reaction'' force. Unlike simple scalar reactions, this Clifford interaction generates \textit{geometric vorticity}. It preserves and enhances structural boundaries (edges) where the feature vector diverges orthogonally from its local context, counteracting the smoothing effects of diffusion.
\end{itemize}

This perspective offers a compelling explanation for the network's effectiveness without FFNs: the architecture evolves raw pixel data into structured semantic representations through a cascade of \textit{geometric pattern formation} steps, mirroring the emergence of complex structures in nature.

\subsection{Hardware-Aware Geometric Approximation}
Our implementation reflects a philosophy of \textit{Pragmatic Platonism}: bridging ideal mathematical structures with hardware realities.

\begin{itemize}
    \item \textbf{Factorized Linear Laplacian:} While the context $\mathcal{C}(H)$ represents a continuous Laplacian operator, we approximate it via a stack of two $3 \times 3$ depth-wise convolutions. This factorization leverages the highly optimized convolution primitives (cuDNN) on modern GPUs, ensuring high throughput while maintaining the mathematical property of iterative diffusion.
    
    \item \textbf{The Cost of Shifts:} Our efficiency analysis reveals that the computational bottleneck lies in the density of geometric sampling (number of shifts), rather than the network depth. Currently, our reliance on standard PyTorch primitives (e.g., \texttt{torch.roll}) incurs memory-bound overheads. However, this also highlights a significant opportunity: since the algorithm is theoretically $\mathcal{O}(N)$, future custom kernel fusion (e.g., via Triton) could perform rolling interactions on-the-fly in registers, potentially doubling the throughput and bridging the gap with mature CNNs.
\end{itemize}

% ----------------------------------------------------------------------
% 6. CONCLUSION
% ----------------------------------------------------------------------
\section{Conclusion and Future Work}

\lettrine[lines=2]{W}e presented the \textbf{Clifford Algebra Network (CAN)}, a vision backbone grounded in algebraic completeness rather than heuristic engineering. By operationalizing the Clifford Geometric Product as a linear-complexity token mixer, we demonstrated that FFNs can be rendered redundant, leading to a new class of highly efficient and geometrically interpretable models. 

By adhering to the mathematical principle of \textbf{algebraic completeness}, we derived a geometric interaction layer that is efficient, robust, and expressive. Our empirical results — particularly the success of the ``No-FFN'' variant — suggest that with the right geometric priors, deep learning models can be significantly simplified, potentially to the point where \textit{geometry is all you need}. The name "Clifford Algebra Network" is deliberate: our current implementation only scratches the surface of what is possible within this rich mathematical framework. We highlight several promising avenues for future research:

\begin{itemize}

    \item \textbf{Scaling to Large-Scale Data:} We plan to validate \modelname{} on ImageNet-1K and ImageNet-21K. A key open question is whether the ``No-FFN'' paradigm maintains its efficacy when the data scale increases by orders of magnitude, or if the geometric interaction requires wider channel mixing to capture the diversity of real-world scenes.
    \item \textbf{Dense Prediction Tasks:} Due to its strict linear complexity $\mathcal{O}(N)$, \modelname{} is theoretically ideal for high-resolution tasks such as semantic segmentation and object detection. Unlike standard ViTs that struggle with quadratic costs at high resolutions, our architecture can process fine-grained feature maps (e.g., $1024 \times 1024$) efficiently, making it a strong candidate for next-generation dense vision backbones.
    \item \textbf{Higher-Order Geometric Products:} Our model utilizes the product of two vectors (Grade 1 $\times$ Grade 1), which generates scalars (Grade 0) and bivectors (Grade 2). Future versions of CAN could explore interactions with higher-order geometric entities, such as the product of a vector and a bivector ($v \cdot (u \wedge w)$). This would allow the network to model more complex volumetric or rotational dynamics, potentially benefiting 3D vision and video understanding.    
    \item \textbf{Optimized Geometric Topology (Decoupled Sampling):} Currently, we share the same shift set $\mathcal{S}$ for both inner and exterior products. However, our ablation study suggests a degree of informational redundancy between the scalar and bivector streams. A promising direction is to decouple the interaction topologies ($\mathcal{S}_{dot} \neq \mathcal{S}_{wedge}$). By assigning distinct receptive fields to the coherent (Inner) and structural (Wedge) pathways — essentially sampling independent diagonals of the interaction matrix — we could maximize the information entropy of the layer without increasing computational cost.
    \item \textbf{Adaptive Geometric Topology:} Currently, our sparse rolling mechanism employs a simple, heuristic set of fixed offsets (e.g., exponential powers), applied uniformly to both the inner (Dot) and exterior (Wedge) product pathways. This assumes that feature coherence and structural contrast operate at identical channel distances, which may not be optimal. Future work could explore \textbf{decoupling these interaction topologies} — finding distinct shift sets tailored for scalar vs. bivector interactions — or introducing \textbf{learnable/adaptive shifts} to dynamically discover the optimal geometric connectivity for specific tasks. This can be viewed as learning the connection coefficients on the feature manifold.
    \item \textbf{Intrinsic Manifold Learning:} Currently, our approach operates on an embedding of the geometric product. Future work could explore intrinsic manifold operations, such as defining metrics based on geodesic distances on the Grassmannian or integrating parallel transport (as explored in RiemannFormer \cite{ji2025riemannformerframeworkattentioncurved}), to further tighten the connection between differential geometry and representation learning.
    \item \textbf{Geometric Attention Augmentation:} Our ``Shifted Generalized Inner Product'' reveals a limitation in standard Self-Attention: the scalar dot product ($\mathbf{q} \cdot \mathbf{k}$) ignores channel topology. We envision extending this concept to Transformers by introducing \textbf{Geometric Multi-Head Attention}, where attention heads are defined not by learned projections, but by cyclic channel shifts. Combined with sparse attention mechanisms (e.g., windowing), this could inject rich geometric inductive biases into ViTs without incurring quadratic costs.
    \item \textbf{Hardware-Aware Optimization:} While our algorithmic complexity is strictly linear, our current implementation relies on standard PyTorch primitives (e.g., \texttt{torch.roll}) which involve memory-bound permutation and broadcasting. We plan to develop custom \textbf{fused CUDA/Triton kernels} to perform the rolling interaction and geometric product on-the-fly within registers. This would eliminate intermediate memory I/O, potentially delivering another $2\times$--$3\times$ wall-clock speedup and bridging the engineering gap with highly optimized convolution libraries (cuDNN).
    \item \textbf{Geometric Multi-Modal Fusion:} Beyond visual backbones, the algebraic structure of \modelname{} offers a universal interface for multi-modal learning (e.g., Vision-Language models). By treating $H$ and $\mathcal{C}$ as heterogeneous modalities (e.g., Image and Text tokens), the Clifford Geometric Product extends standard \textit{Cross-Attention} (scalar alignment) to include \textbf{Cross-Modal Bivectors} (wedge product). This implies a new fusion paradigm where modalities are not just aligned, but geometrically \textit{composed} into higher-order semantic planes, potentially offering richer representations for generative tasks.
    \item \textbf{Data-Dependent Manifold Geometry:} Currently, our Laplacian context $\mathcal{C}(H)$ relies on static convolution kernels, implying a fixed geometric prior for all images. A compelling direction, inspired by \textbf{RiemannFormer} \cite{ji2025riemannformerframeworkattentioncurved}, is to make the metric structure \textit{dynamic}. By formulating the LBO as a function of the input features themselves (e.g., via dynamic convolutions or local attention), we could model a coupled system where ``matter tells space how to curve'' (features define the metric) and ``space tells matter how to move'' (the metric guides the Clifford flow).
    \item \textbf{Symplectic Geometric Flow:} The prominence of the wedge product ($H \wedge \mathcal{C}$) suggests a deep connection to \textbf{Symplectic Geometry}, where dynamics are governed by a skew-symmetric symplectic form. While our current task (classification) requires a dissipative system to collapse information into labels, future work could explore \textbf{Symplectic CliffordNets}. By enforcing symplectic integrators (energy-preserving flows), this architecture could be naturally adapted for tasks requiring conservation laws, such as \textbf{physical dynamics simulation} or \textbf{video generation}, effectively bridging the gap between computer vision and Hamiltonian mechanics.
\end{itemize}

%Bibliography
\bibliographystyle{unsrt} 
\bibliography{can}  

\end{document}